\documentclass[11pt,a4paper]{article}
\pdfoutput=1

\usepackage[hyperref]{acl2019}
\hypersetup{
    colorlinks=true,
    linkcolor=blue,
    filecolor=red,
    urlcolor=magenta,
    breaklinks=true,
}
\usepackage{breakurl}
\usepackage{acl2019}
\usepackage{times}
\usepackage{latexsym}

\usepackage{url}

\aclfinalcopy %
\def\aclpaperid{1553} %

\usepackage{amsmath}
\usepackage{amssymb}
\usepackage{amsthm}
\usepackage{booktabs}
\usepackage{color}
\usepackage{graphicx}
\usepackage{multirow}
\usepackage{subfigure}
\usepackage{verbatim}
\usepackage{tipa}
\usepackage{CJKutf8}

\title{Learning to Discover, Ground and Use Words\\with Segmental Neural Language Models}
\author{Kazuya Kawakami$^{\spadesuit\clubsuit}$
		~ Chris Dyer$^{\clubsuit}$
		~ Phil Blunsom$^{\spadesuit\clubsuit}$\\
$^{\spadesuit}$Department of Computer Science, University of Oxford, Oxford, UK \\
$^{\clubsuit}$DeepMind, London, UK \\\
{\small \tt \{kawakamik,cdyer,pblunsom\}@google.com}
}

\begin{document}
\maketitle

\begin{abstract}
We propose a segmental neural language model that combines the generalization power of neural networks with the ability to discover word-like units that are latent in unsegmented character sequences. In contrast to previous segmentation models that treat word segmentation as an isolated task, our model unifies word discovery, learning how words fit together to form sentences, and, by conditioning the model on visual context, how words' meanings ground in representations of non-linguistic modalities. Experiments show that the unconditional model learns predictive distributions better than character LSTM models, discovers words competitively with nonparametric Bayesian word segmentation models, and that modeling language conditional on visual context improves performance on both.
\end{abstract}

\section{Introduction}
How infants discover words that make up their first language is a long-standing question in developmental psychology~\citep{saffran1996statistical}. Machine learning has contributed much to this discussion by showing that predictive models of language are capable of inferring the existence of word boundaries solely based on statistical properties of the input~\citep{elman1990finding,brent1996distributional,goldwater2009bayesian}. However, there are two serious limitations of current models of word learning in the context of the broader problem of language acquisition. First, language acquisition involves not only learning what words there are (``the lexicon''), but also how they fit together (``the grammar''). Unfortunately, the best language models, measured in terms of their ability to predict language (i.e., those which seem acquire grammar best), segment quite poorly~\citep{chung2016hierarchical,wang:2017,kdr2018revisiting}, while the strongest models in terms of word segmentation~\citep{goldwater2009bayesian,berg2010painless} do not adequately account for the long-range dependencies that are manifest in language and that are easily captured by recurrent neural networks~\citep{mikolov2010recurrent}. Second, word learning involves not only discovering what words exist and how they fit together grammatically, but also determining their non-linguistic referents, that is, their grounding. The work that has looked at modeling acquisition of grounded language from character sequences---usually in the context of linking words to a visually experienced environment---has either explicitly avoided modeling word units~\citep{gelderloos:2016} or relied on high-level representations of visual context that overly simplify the richness and ambiguity of the visual signal~\citep{johnson:2010,rasanen:2015}.

In this paper, we introduce a single model that discovers words, learns how they fit together (not just locally, but across a complete sentence), and grounds them in learned representations of naturalistic non-linguistic visual contexts. We argue that such a unified model is preferable to a pipeline model of language acquisition (e.g., a model where words are learned by one character-aware model, and then a full-sentence grammar is acquired by a second language model using the words predicted by the first). Our preference for the unified model may be expressed in terms of basic notions of simplicity (we require one model rather than two), and in terms of the Continuity Hypothesis of \citet{pinker:1984}, which argues that we should assume, absent strong evidence to the contrary, that children have the same cognitive systems as adults, and differences are due to them having set their parameters differently/immaturely.

Our model depends crucially on two components. The first is, as mentioned, a lexical memory. This lexicon stores pairs of a vector (key) and a string (value) the strings in the lexicon are contiguous sequences of characters encountered in the training data; and the vectors are randomly initialized and learned during training. The second component is a regularizer~(\S\ref{sec:reg}) that prevents the model from overfitting to the training data by overusing the lexicon to account for the training data.\footnote{Since the lexical memory stores strings that appear in the training data, each sentence could, in principle, be generated as a single lexical unit, thus the model could fit the training data perfectly while generalizing poorly. The regularizer penalizes based on the expectation of the powered length of each segment, preventing this degenerate solution from being optimal.}

Our evaluation (\S\ref{sec:dataset}--\S\ref{sec:results}) looks at both language modeling performance and the quality of the induced segmentations, in both unconditional (sequence-only) contexts and when conditioning on a related image. First, we look at the segmentations induced by our model. We find that these correspond closely to human intuitions about word segments, competitive with the best existing models for unsupervised word discovery. Importantly, these segments are obtained in models whose hyperparameters are tuned to optimize validation (held-out) likelihood, whereas tuning the hyperparameters of our benchmark models using held-out likelihood produces poor segmentations. Second, we confirm findings~\citep{kawakami2017learning,mielke:2018} that show that word segmentation information leads to better language models compared to pure character models. However, in contrast to previous work, we realize this performance improvement without having to observe the segment boundaries. Thus, our model may be applied straightforwardly to Chinese, where word boundaries are not part of the orthography.

Ablation studies demonstrate that both the lexicon and the regularizer are crucial for good performance, particularly in word segmentation---removing either or both significantly harms performance. In a final experiment, we learn to model language that describes images, and we find that conditioning on visual context improves segmentation performance in our model (compared to the performance when the model does not have access to the image). On the other hand, in a baseline model that predicts boundaries based on entropy spikes in a character-LSTM, making the image available to the model has no impact on the quality of the induced segments, demonstrating again the value of explicitly including a word lexicon in the language model.

\section{Model}\label{sec:model}
\begin{figure*}[htbp]
    \begin{center}
        \includegraphics[width=0.95\linewidth]{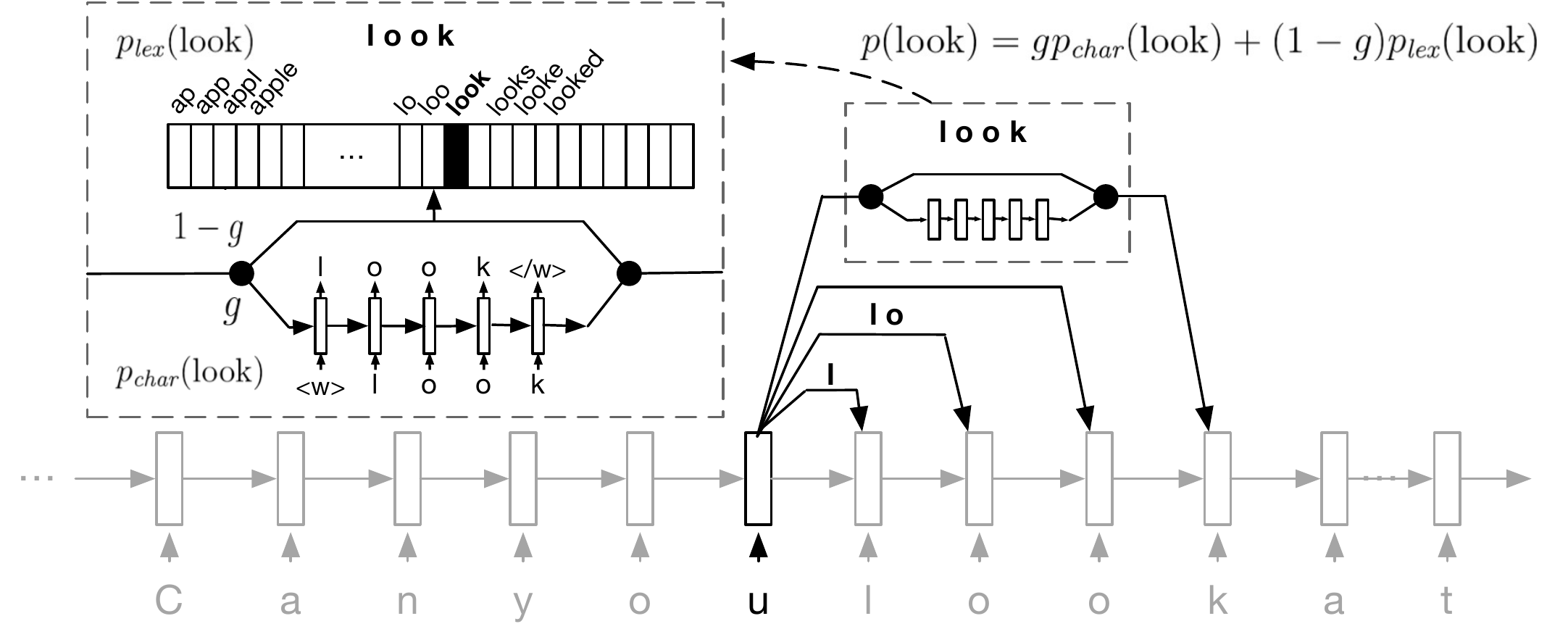}
    \end{center}
    \caption{Fragment of the segmental neural language model while evaluating the marginal likelihood of a sequence. At the indicated time, the model has generated the sequence \emph{Canyou}, and four possible continuations are shown.}
    \label{fig:model}
\end{figure*}

We now describe the segmental neural language model (SNLM). Refer to Figure~\ref{fig:model} for an illustration. The SNLM generates a character sequence $\boldsymbol{x}=x_{1}, \ldots, x_{n}$, where each $x_i$ is a character in a finite character set $\Sigma$. Each sequence $\boldsymbol{x}$ is the concatenation of a sequence of segments $\underline{\boldsymbol{s}} = \boldsymbol{s}_1, \ldots, \boldsymbol{s}_{|\underline{\boldsymbol{s}}|}$ where $|\underline{\boldsymbol{s}}| \le n$ measures the length of the sequence in segments and each segment $\boldsymbol{s}_i \in \Sigma^+$ is a sequence of characters, $s_{i,1},\ldots,s_{i,|\boldsymbol{s}_i|}$. Intuitively, each $\boldsymbol{s}_i$ corresponds to one word. Let $\pi(\boldsymbol{s}_1, \ldots, \boldsymbol{s}_{i})$ represent the concatenation of the characters of the segments $\boldsymbol{s}_1$ to $\boldsymbol{s}_i$, discarding segmentation information; thus $\boldsymbol{x}=\pi(\underline{\boldsymbol{s}})$. For example if $\boldsymbol{x} = \texttt{anapple}$, the underlying segmentation might be $\underline{\boldsymbol{s}}=\texttt{an apple}$ (with $\boldsymbol{s}_1=\texttt{an}$ and $\boldsymbol{s}_2=\texttt{apple}$), or $\underline{\boldsymbol{s}}=\texttt{a nap ple}$, or any of the $2^{|\boldsymbol{x}|-1}$ segmentation possibilities for $\boldsymbol{x}$.

The SNLM defines the distribution over $\boldsymbol{x}$ as the marginal distribution over all segmentations that give rise to $\boldsymbol{x}$, i.e.,
\begin{align}
    p(\boldsymbol{x}) = \sum_{\underline{\boldsymbol{s}} : \pi(\underline{\boldsymbol{s}}) = \boldsymbol{x}} p(\underline{\boldsymbol{s}}). \label{eq:marginal}
\end{align}
To define the probability of $p(\underline{\boldsymbol{s}})$, we use the chain rule, rewriting this in terms of a product of the series of conditional probabilities, $p(\boldsymbol{s}_t \mid \underline{\boldsymbol{s}}_{<t})$. The process stops when a special end-sequence segment $\langle /\textsc{s} \rangle$ is generated. To ensure that the summation in Eq.~\ref{eq:marginal} is tractable, we assume the following:
\begin{align}
    p(\boldsymbol{s}_t \mid \underline{\boldsymbol{s}}_{<t}) &\approx p(\boldsymbol{s}_t \mid \pi(\underline{\boldsymbol{s}}_{<t})) = p(\boldsymbol{s}_t \mid \boldsymbol{x}_{<t}),
\end{align}
which amounts to a conditional semi-Markov assumption---i.e., non-Markovian generation happens inside each segment, but the segment generation probability does not depend on memory of the previous segmentation decisions, only upon the sequence of characters $\pi(\underline{\boldsymbol{s}}_{<t})$ corresponding to the prefix character sequence $\boldsymbol{x}_{<t}$. This assumption has been employed in a number of related models to permit the use of LSTMs to represent rich history while retaining the convenience of dynamic programming inference algorithms~\citep{wang:2017,ling2016latent,graves2012sequence}.

\subsection{Segment generation}
We model $p(\boldsymbol{s}_t \mid \boldsymbol{x}_{<t})$ as a mixture of two models, one that generates the segment using a sequence model and the other that generates multi-character sequences as a single event. Both are conditional on a common representation of the history, as is the mixture proportion.

\paragraph{Representing history}
To represent $\boldsymbol{x}_{<t}$, we use an LSTM encoder to read the sequence of characters, where each character type $\sigma \in \Sigma$ has a learned vector embedding $\mathbf{v}_{\sigma}$. Thus the history representation at time $t$ is $\mathbf{h}_t = \mathrm{LSTM}_{\textit{enc}}(\mathbf{v}_{x_1}, \ldots, \mathbf{v}_{x_t})$.
This corresponds to the standard history representation for a character-level language model, although in general, we assume that our modelled data is not delimited by whitespace.

\paragraph{Character-by-character generation}
The first component model, $p_{\textit{char}}(\boldsymbol{s}_{t}\mid \mathbf{h}_t)$, generates $\boldsymbol{s}_t$ by sampling a sequence of characters from a LSTM language model over $\Sigma$ and a two extra special symbols, an end-of-word symbol $\langle \textsc{/w} \rangle \notin \Sigma$ and the end-of-sequence symbol $\langle /\textsc{s} \rangle$ discussed above. The initial state of the LSTM is a learned transformation of $\mathbf{h}_t$, the initial cell is $\mathbf{0}$, and different parameters than the history encoding LSTM are used. During generation, each letter that is sampled (i.e., each $s_{t,i}$) is fed back into the LSTM in the usual way and the probability of the character sequence decomposes according to the chain rule. The end-of-sequence symbol can never be generated in the initial position.

\paragraph{Lexical generation}
The second component model, $p_{\textit{lex}}(\boldsymbol{s}_t \mid \mathbf{h}_t)$, samples full segments from lexical memory. Lexical memory is a key-value memory containing $M$ entries, where each key, $\mathbf{k}_i$, a vector, is associated with a value $\boldsymbol{v}_i \in \Sigma^+$. The generation probability of $\boldsymbol{s}_t$ is defined as
\begin{align*}
    \mathbf{h}'_t &= \textrm{MLP}(\mathbf{h}_t) \\
    \mathbf{m} &= \mathrm{softmax}(\mathbf{Kh}'_t + \mathbf{b}) \\
    p_{\textit{lex}}(\boldsymbol{s}_t \mid \mathbf{h}_t) &= \sum_{i=1}^M m_i[\boldsymbol{v}_i = \boldsymbol{s}_t],
\end{align*}
where $[\boldsymbol{v}_i = \boldsymbol{s}_t]$ is 1 if the $i$th value in memory is $\boldsymbol{s}_t$ and 0 otherwise, and $\mathbf{K}$ is a matrix obtained by stacking the $\mathbf{k}_i^{\top}$'s. This generation process assigns zero probability to most strings, but the alternate character model can generate all of $\Sigma^+$.

In this work, we fix the $\boldsymbol{v}_i$'s to be subsequences of at least length 2, and up to a maximum length $L$ that are observed at least $F$ times in the training data. These values are tuned as hyperparameters (See Appendix~\ref{sec:config} for details of the experiments).

\paragraph{Mixture proportion}
The mixture proportion, $g_t$, determines how likely the character generator is to be used at time $t$ (the lexicon is used with probability $1-g_t$). It is defined by as $g_t = \sigma(\mathrm{MLP}(\mathbf{h}_t))$.

\paragraph{Total segment probability}
The total generation probability of $\boldsymbol{s}_t$ is thus
\begin{align*}
p(\boldsymbol{s}_{t}\mid \boldsymbol{x}_{<t}) &= g_{t} p_{\textit{char}}(\boldsymbol{s}_{t}\mid \mathbf{h}_t) + \\
&\qquad \ \ (1- g_{t})p_{\textit{lex}}(\boldsymbol{s}_{t}\mid \mathbf{h}_t).
\end{align*}

\section{Inference}\label{sec:inference}
We are interested in two inference questions: first, given a sequence $\boldsymbol{x}$, evaluate its (log) marginal likelihood; second, given $\boldsymbol{x}$, find the most likely decomposition into segments $\underline{\boldsymbol{s}}^*$. %

\paragraph{Marginal likelihood}
To efficiently compute the marginal likelihood, we use a variant of the forward algorithm for semi-Markov models~\citep{yu:2010}, which incrementally computes a sequence of probabilities, $\alpha_i$, where $\alpha_i$ is the marginal likelihood of generating $\boldsymbol{x}_{\le i}$ and concluding a segment at time $i$. Although there are an exponential number of segmentations of $\boldsymbol{x}$, these values can be computed using $O(|\boldsymbol{x}|)$ space and $O(|\boldsymbol{x}|^2)$ time as:
\begin{align}
\alpha_0 = 1, \qquad
\alpha_{t} = \sum_{j=t-L}^{t-1} \alpha_{j} p(\boldsymbol{s} = \boldsymbol{x}_{j:t}\mid \boldsymbol{x}_{<j}). \label{eq:forward}
\end{align}
By letting $x_{t+1} = \langle /\textsc{s} \rangle$, then $p(\boldsymbol{x}) = \alpha_{t+1}$.

\paragraph{Most probable segmentation}
The most probable segmentation of a sequence $\boldsymbol{x}$ can be computed by replacing the summation with a $\max$ operator in Eq.~\ref{eq:forward} and maintaining backpointers.

\section{Expected length regularization}\label{sec:reg}
When the lexical memory contains all the substrings in the training data, the model easily overfits by copying the longest continuation from the memory. To prevent overfitting, we introduce a regularizer that penalizes based on the expectation of the exponentiated (by a hyperparameter $\beta$) length of each segment:
\begin{align*}
    R(\boldsymbol{x}, \beta) = \sum_{\underline{\boldsymbol{s}} : \pi(\underline{\boldsymbol{s}})=\boldsymbol{x}} p(\underline{\boldsymbol{s}} \mid \boldsymbol{x}) \sum_{\boldsymbol{s} \in \underline{\boldsymbol{s}}} |\boldsymbol{s}|^{\beta}.
\end{align*}
This can be understood as a regularizer based on the double exponential prior identified to be effective in previous work~\citep{liang2009online,berg2010painless}. This expectation is a differentiable function of the model parameters. Because of the linearity of the penalty across segments, it can be computed efficiently using the above dynamic programming algorithm under the expectation semiring~\citep{eisner2002parameter}. This is particularly efficient since the expectation semiring jointly computes the expectation and marginal likelihood in a single forward pass. For more details about computing gradients of expectations under distributions over structured objects with dynamic programs and semirings, see~\citet{li2009first}.

\subsection{Training Objective}
The model parameters are trained by minimizing the penalized log likelihood of a training corpus $\mathcal{D}$ of unsegmented sentences,
\begin{align*}
	\mathcal{L} = \sum_{\boldsymbol{x} \in \mathcal{D}}[-\log p(\boldsymbol{x}) + \lambda R(\boldsymbol{x}, \beta)].
\end{align*}

\section{Datasets}\label{sec:dataset}
We evaluate our model on both English and Chinese segmentation. For both languages, we used standard datasets for word segmentation and language modeling. We also use MS-COCO to evaluate how the model can leverage conditioning context information. For all datasets, we used train, validation and test splits.\footnote{The data and splits used are available at \url{https://s3.eu-west-2.amazonaws.com/k-kawakami/seg.zip}.} Since our model assumes a closed character set, we removed validation and test samples which contain characters that do not appear in the training set. In the English corpora, whitespace characters are removed. In Chinese, they are not present to begin with. Refer to Appendix~\ref{sec:app-stats} for dataset statistics.

\subsection{English}
\paragraph{Brent Corpus} The Brent corpus is a standard corpus used in statistical modeling of child language acquisition~\citep{brent1999efficient,venkataraman2001statistical}.\footnote{\url{https://childes.talkbank.org/derived}} The corpus contains transcriptions of utterances directed at 13- to 23-month-old children. The corpus has two variants: an orthographic one (\textbf{BR-text}) and a phonemic one (\textbf{BR-phono}), where each character corresponds to a single English phoneme. As the Brent corpus does not have a standard train and test split, and we want to tune the parameters by measuring the fit to held-out data, we used the first 80\% of the utterances for training and the next 10\% for validation and the rest for test.

\paragraph{English Penn Treebank (PTB)} We use the commonly used version of the PTB prepared by \citet{mikolov2010recurrent}. However, since we removed space symbols from the corpus, our cross entropy results cannot be compared to those usually reported on this dataset.

\subsection{Chinese}
Since Chinese orthography does not mark spaces between words, there have been a number of efforts to annotate word boundaries. We evaluate against two corpora that have been manually segmented according different segmentation standards.
\paragraph{Beijing University Corpus (PKU)} The Beijing University Corpus was one of the corpora used for the International Chinese Word Segmentation Bakeoff~\citep{emerson2005second}.
\paragraph{Chinese Penn Treebank (CTB)} We use the Penn Chinese Treebank Version 5.1 \citep{xue2005penn}. It generally has a coarser segmentation than PKU (e.g., in CTB a full name, consisting of a given name and family name, is a single token), and it is a larger corpus.

\subsection{Image Caption Dataset}
To assess whether jointly learning about meanings of words from non-linguistic context affects segmentation performance, we use image and caption pairs from the COCO caption dataset~\cite{lin2014microsoft}. We use 10,000 examples for both training and testing and we only use one reference per image. The images are used to be conditional context to predict captions. Refer to Appendix~\ref{sec:app-caption} for the dataset construction process.

\section{Experiments}\label{sec:experiments}
We compare our model to benchmark Bayesian models, which are currently the best known unsupervised word discovery models, as well as to a simple deterministic segmentation criterion based on surprisal peaks~\citep{elman1990finding} on language modeling and segmentation performance. Although the Bayeisan models are shown to able to discover plausible word-like units, we found that a set of hyperparameters that provides best performance with such model on language modeling does not produce good structures as reported in previous works. This is problematic since there is no objective criteria to find hyperparameters in fully unsupervised manner when the model is applied to completely unknown languages or domains. Thus, our experiments are designed to assess how well the models infers word segmentations of unsegmented inputs when they are trained and tuned to maximize the likelihood of the held-out text.

\paragraph{DP/HDP Benchmarks}
Among the most effective existing word segmentation models are those based on hierarchical Dirichlet process (HDP) models~\citep{goldwater2009bayesian,teh2006hierarchical} and hierarchical Pitman--Yor processes~\citep{mochihashi2009bayesian}. As a representative of these, we use a simple bigram HDP model:
\begin{align*}
    \theta_{\cdot} &\sim \mathrm{DP}(\alpha_0, p_0) \\
    \theta_{\cdot \mid \boldsymbol{s}} &\sim \mathrm{DP}(\alpha_1, \theta_{\cdot}) \qquad \forall \boldsymbol{s} \in \Sigma^* \\
    \boldsymbol{s}_{t+1} \mid \boldsymbol{s}_t &\sim \mathrm{Categorical}(\theta_{\cdot \mid \boldsymbol{s}_t}).
\end{align*}
The base distribution, $p_0$, is defined over strings in $\Sigma^* \cup \{ \langle /\textsc{s} \rangle \}$ by deciding with a specified probability to end the utterance, a geometric length model, and a uniform probability over $\Sigma$ at a each position. Intuitively, it captures the preference for having short words in the lexicon. In addition to the HDP model, we also evaluate a simpler single Dirichlet process (DP) version of the model, in which the $\boldsymbol{s}_t$'s are generated directly as draws from $\mathrm{Categorical}(\theta_{\cdot})$. We use an empirical Bayesian approach to select hyperparameters based on the likelihood assigned by the inferred posterior to a held-out validation set. Refer to Appendix~\ref{app:inference} for details on inference.

\paragraph{Deterministic Baselines}
Incremental word segmentation is inherently ambiguous (e.g., the letters \emph{the} might be a single word, or they might be the beginning of the longer word \emph{theater}). Nevertheless, several deterministic functions of prefixes have been proposed in the literature as strategies for discovering rudimentary word-like units hypothesized for being useful for bootstrapping the lexical acquisition process or for improving a model's predictive accuracy. These range from surprisal criteria~\citep{elman1990finding} to sophisticated language models that switch between models that capture intra- and inter-word dynamics based on deterministic functions of prefixes of characters~\citep{chung2016hierarchical,shen2017neural}.

In our experiments, we also include such deterministic segmentation results using (1) the surprisal criterion of \citet{elman1990finding} and (2) a two-level hierarchical multiscale LSTM~\citep{chung2016hierarchical}, which has been shown to predict boundaries in whitespace-containing character sequences at positions corresponding to word boundaries. As with all experiments in this paper, the BR-corpora for this experiment do not contain spaces.

\paragraph{SNLM Model configurations and Evaluation}
LSTMs had 512 hidden units with parameters learned using the Adam update rule~\citep{kingma2014adam}. We evaluated our models with bits-per-character (bpc) and segmentation accuracy~\citep{brent1999efficient,venkataraman2001statistical,goldwater2009bayesian}. Refer to Appendices~\ref{sec:config}--\ref{sec:metrics} for details of model configurations and evaluation metrics.

For the image caption dataset, we extend the model with a standard attention mechanism in the backbone LSTM ($\mathrm{LSTM}_{\textit{enc}}$) to incorporate image context. For every character-input, the model calculates attentions over image features and use them to predict the next characters. As for image representations, we use features from the last convolution layer of a pre-trained VGG19 model~\cite{simonyan14verydeep}.

\section{Results}\label{sec:results}
In this section, we first do a careful comparison of segmentation performance on the phonemic Brent corpus (BR-phono) across several different segmentation baselines, and we find that our model obtains competitive segmentation performance. Additionally, ablation experiments demonstrate that both lexical memory and the proposed expected length regularization are necessary for inferring good segmentations. We then show that also on other corpora, we likewise obtain segmentations better than baseline models. Finally, we also show that our model has superior performance, in terms of held-out perplexity, compared to a character-level LSTM language model. Thus, overall, our results show that we can obtain good segmentations on a variety of tasks, while still having very good language modeling performance.

\paragraph{Word Segmentation (BR-phono)}
Table~\ref{tb:result-phono} summarizes the segmentation results on the widely used BR-phono corpus, comparing it to a variety of baselines.
\textbf{Unigram DP}, \textbf{Bigram HDP}, \textbf{LSTM suprisal} and \textbf{HMLSTM} refer to the benchmark models explained in \S\ref{sec:experiments}. 
The ablated versions of our model show that without the lexicon ($-$memory), without the expected length penalty ($-$length), and without either, our model fails to discover good segmentations. Furthermore, we draw attention to the difference in the performance of the HDP and DP models when using subjective settings of the hyperparameters and the empirical settings (likelihood). Finally, the deterministic baselines are interesting in two ways. First, LSTM surprisal is a remarkably good heuristic for segmenting text (although we will see below that its performance is much less good on other datasets). Second, despite careful tuning, the HMLSTM of \citet{chung2016hierarchical} fails to discover good segments, although in their paper they show that when spaces are present between, HMLSTMs learn to switch between their internal models in response to them.

Furthermore, the priors used in the DP/HDP models were tuned to maximize the likelihood assigned to the validation set by the inferred posterior predictive distribution, in contrast to previous papers which either set them subjectively or inferred them~\cite{johnson2009improving}. For example, the DP and HDP model with subjective priors obtained 53.8 and 72.3 F1 scores, respectively~\citep{goldwater2009bayesian}. However, when the hyperparameters are set to maximize held-out likelihood, this drops obtained 56.1 and 56.9. Another result on this dataset is the feature unigram model of \citet{berg2010painless}, which obtains an 88.0 F1 score with hand-crafted features and by selecting the regularization strength to optimize segmentation performance. Once the features are removed, the model achieved a 71.5 F1 score when it is tuned on segmentation performance and only 11.5 when it is tuned on held-out likelihood.

\begin{table}[ht]
    \begin{center}%
    \begin{tabular}{l@{\hskip 4pt}c@{\hskip 4pt}c@{\hskip 4pt}c@{\hskip 4pt}}
    \toprule
     & P & R & F1\\
    \midrule
    LSTM surprisal~\citep{elman1990finding} & 54.5 & 55.5 & 55.0\\
    HMLSTM~\citep{chung2016hierarchical} & 8.1 & 13.3 & 10.1\\
    \midrule[0.2ex]
    Unigram DP & 63.3 & 50.4 & 56.1\\
    Bigram HDP & 53.0 & 61.4 & 56.9\\
    SNLM ($-$memory, $-$length) & 54.3 & 34.9 & 42.5\\
    SNLM ($+$memory, $-$length) & 52.4 & 36.8 & 43.3\\
    SNLM ($-$memory, $+$length) & 57.6 & 43.4 & 49.5\\
    SNLM ($+$memory, $+$length)& \textbf{81.3} & \textbf{77.5} & \textbf{79.3}\\
    \bottomrule[0.3ex]
    \end{tabular}
    \end{center}
        \caption{Summary of segmentation performance on phoneme version of the Brent Corpus (\textbf{BR-phono}).
        \label{tb:result-phono}}
\end{table}

\paragraph{Word Segmentation (other corpora)} Table~\ref{tb:result-seg} summarizes results on the BR-text (orthographic Brent corpus) and Chinese corpora. As in the previous section, all the models were trained to maximize held-out likelihood. Here we observe a similar pattern, with the SNLM outperforming the baseline models, despite the tasks being quite different from each other and from the BR-phono task. 

\begin{table}[ht]
    \begin{center}%
    \begin{tabular}{clccc}
    \toprule
    & & P & R & F1\\
    \midrule
    \multirow{4}{*}{BR-text} 
    & LSTM surprisal & 36.4 & 49.0 & 41.7\\
    & Unigram DP   & 64.9 & 55.7 & 60.0\\
    & Bigram HDP  & 52.5 & 63.1 & 57.3\\
    & SNLM & \textbf{68.7} & \textbf{78.9} & \textbf{73.5}\\
    \midrule
    \multirow{4}{*}{PTB} 
    & LSTM surprisal & 27.3 & 36.5 & 31.2\\
    & Unigram DP   & 51.0 & 49.1 & 50.0\\
    & Bigram HDP  & 34.8 & 47.3 & 40.1\\
    & SNLM & \textbf{54.1} & \textbf{60.1} & \textbf{56.9}\\
    \midrule
    \multirow{4}{*}{CTB} 
    & LSTM surprisal & 41.6 & 25.6 & 31.7\\
    & Unigram DP   & 61.8 & 49.6 & 55.0\\
    & Bigram HDP  & 67.3 & 67.7 & 67.5\\
    & SNLM & \textbf{78.1} & \textbf{81.5} & \textbf{79.8}\\
    \midrule
    \multirow{4}{*}{PKU}
    & LSTM surprisal & 38.1 & 23.0 & 28.7\\
    & Unigram DP  & 60.2 & 48.2 & 53.6\\
    & Bigram HDP  & 66.8 & 67.1 & 66.9\\
    & SNLM & \textbf{75.0} & \textbf{71.2} & \textbf{73.1}\\
    \bottomrule[0.3ex]
    \end{tabular}
    \end{center}
        \caption{Summary of segmentation performance on other corpora.\label{tb:result-seg}}
\end{table}

\paragraph{Word Segmentation Qualitative Analysis}
We show some representative examples of segmentations inferred by various models on the BR-text and PKU corpora in Table~\ref{tb:quality}. As reported in \citet{goldwater2009bayesian}, we observe that the DP models tend to undersegment, keep long frequent sequences together (e.g., they failed to separate articles). HDPs do successfully prevent oversegmentation; however, we find that when trained to optimize held-out likelihood, they often insert unnecessary boundaries between words, such as \textit{yo u}. Our model's performance is better, but it likewise shows a tendency to oversegment. Interestingly, we can observe a tendency tends to put boundaries between morphemes in morphologically complex lexical items such as \textit{dumpty 's}, and \textit{go ing}. Since morphemes are the minimal units that carry meaning in language, this segmentation, while incorrect, is at least plasuible. Turning to the Chinese examples, we see that both baseline models fail to discover basic words such as \begin{CJK}{UTF8}{gbsn}山间\end{CJK} (mountain) and \begin{CJK}{UTF8}{gbsn}人们\end{CJK}~(human).

Finally, we observe that none of the models successfully segment dates or numbers containing multiple digits (all oversegment). Since number types tend to be rare, they are usually not in the lexicon, meaning our model (and the H/DP baselines) must generate them as character sequences.

\begin{table*}[ht]
    \begin{center}\footnotesize
    \begin{tabular}{cll}
    \toprule
    &  & Examples \\
    \midrule
    \multirow{9}{*}{BR-text}
    \\[-1.7ex]
    & Reference & are you going to make him pretty this morning\\
    & Unigram DP  & areyou goingto makehim pretty this morning\\
    & Bigram HDP & areyou go ingto make him p retty this mo rn ing\\
    & SNLM & are you go ing to make him pretty this morning\\
    \\[-2.0ex] \cline{2-3} \\[-1.5ex]
    & Reference & would you like to do humpty dumpty's button\\
    & Unigram DP & wouldyoul iketo do humpty dumpty 's button\\
    & Bigram HDP & would youlike to do humptyd umpty 's butt on\\
    & SNLM & would you like to do humpty dumpty 's button\\
    
    \midrule
    \multirow{9}{*}{PKU}
    \\[-1.7ex]
    & Reference & \begin{CJK}{UTF8}{gbsn}笑声\end{CJK}$\ \ $\begin{CJK}{UTF8}{gbsn}、\end{CJK}$\ \ $\begin{CJK}{UTF8}{gbsn}掌声\end{CJK}$\ \ $\begin{CJK}{UTF8}{gbsn}、\end{CJK}$\ \ $\begin{CJK}{UTF8}{gbsn}欢呼声\end{CJK}$\ \ $\begin{CJK}{UTF8}{gbsn}，\end{CJK}$\ \ $\begin{CJK}{UTF8}{gbsn}在\end{CJK}$\ \ $\begin{CJK}{UTF8}{gbsn}山间\end{CJK}$\ \ $\begin{CJK}{UTF8}{gbsn}回荡\end{CJK}$\ \ $\begin{CJK}{UTF8}{gbsn}，\end{CJK}$\ \ $\begin{CJK}{UTF8}{gbsn}勾\end{CJK}$\ \ $\begin{CJK}{UTF8}{gbsn}起\end{CJK}$\ \ $\begin{CJK}{UTF8}{gbsn}了\end{CJK}$\ \ $\begin{CJK}{UTF8}{gbsn}人们\end{CJK}$\ \ $\begin{CJK}{UTF8}{gbsn}对\end{CJK}$\ \ $\begin{CJK}{UTF8}{gbsn}往事\end{CJK}$\ \ $\begin{CJK}{UTF8}{gbsn}的\end{CJK}$\ \ $\begin{CJK}{UTF8}{gbsn}回忆\end{CJK}$\ \ $\begin{CJK}{UTF8}{gbsn}。\end{CJK}\\
    & Unigram DP & \begin{CJK}{UTF8}{gbsn}笑声\end{CJK}$\ \ $\begin{CJK}{UTF8}{gbsn}、\end{CJK}$\ \ $\begin{CJK}{UTF8}{gbsn}掌声\end{CJK}$\ \ $\begin{CJK}{UTF8}{gbsn}、\end{CJK}$\ \ $\begin{CJK}{UTF8}{gbsn}欢呼\end{CJK}$\ \ $\begin{CJK}{UTF8}{gbsn}声\end{CJK}$\ \ $\begin{CJK}{UTF8}{gbsn}，在\end{CJK}$\ \ $\begin{CJK}{UTF8}{gbsn}山\end{CJK}$\ \ $\begin{CJK}{UTF8}{gbsn}间\end{CJK}$\ \ $\begin{CJK}{UTF8}{gbsn}回荡\end{CJK}$\ \ $\begin{CJK}{UTF8}{gbsn}，\end{CJK}$\ \ $\begin{CJK}{UTF8}{gbsn}勾\end{CJK}$\ \ $\begin{CJK}{UTF8}{gbsn}起了\end{CJK}$\ \ $\begin{CJK}{UTF8}{gbsn}人们对\end{CJK}$\ \ $\begin{CJK}{UTF8}{gbsn}往事\end{CJK}$\ \ $\begin{CJK}{UTF8}{gbsn}的\end{CJK}$\ \ $\begin{CJK}{UTF8}{gbsn}回忆\end{CJK}$\ \ $\begin{CJK}{UTF8}{gbsn}。\end{CJK}\\
    & Bigram HDP & \begin{CJK}{UTF8}{gbsn}笑\end{CJK}$\ \ $\begin{CJK}{UTF8}{gbsn}声、\end{CJK}$\ \ $\begin{CJK}{UTF8}{gbsn}掌声\end{CJK}$\ \ $\begin{CJK}{UTF8}{gbsn}、\end{CJK}$\ \ $\begin{CJK}{UTF8}{gbsn}欢\end{CJK}$\ \ $\begin{CJK}{UTF8}{gbsn}呼声\end{CJK}$\ \ $\begin{CJK}{UTF8}{gbsn}，在\end{CJK}$\ \ $\begin{CJK}{UTF8}{gbsn}山\end{CJK}$\ \ $\begin{CJK}{UTF8}{gbsn}间\end{CJK}$\ \ $\begin{CJK}{UTF8}{gbsn}回\end{CJK}$\ \ $\begin{CJK}{UTF8}{gbsn}荡，\end{CJK}$\ \ $\begin{CJK}{UTF8}{gbsn}勾\end{CJK}$\ \ $\begin{CJK}{UTF8}{gbsn}起了\end{CJK}$\ \ $\begin{CJK}{UTF8}{gbsn}人\end{CJK}$\ \ $\begin{CJK}{UTF8}{gbsn}们对\end{CJK}$\ \ $\begin{CJK}{UTF8}{gbsn}往事\end{CJK}$\ \ $\begin{CJK}{UTF8}{gbsn}的\end{CJK}$\ \ $\begin{CJK}{UTF8}{gbsn}回忆\end{CJK}$\ \ $\begin{CJK}{UTF8}{gbsn}。\end{CJK}\\
    & SNLM & \begin{CJK}{UTF8}{gbsn}笑声、\end{CJK}$\ \ $\begin{CJK}{UTF8}{gbsn}掌声\end{CJK}$\ \ $\begin{CJK}{UTF8}{gbsn}、\end{CJK}$\ \ $\begin{CJK}{UTF8}{gbsn}欢呼声\end{CJK}$\ \ $\begin{CJK}{UTF8}{gbsn}，\end{CJK}$\ \ $\begin{CJK}{UTF8}{gbsn}在\end{CJK}$\ \ $\begin{CJK}{UTF8}{gbsn}山间\end{CJK}$\ \ $\begin{CJK}{UTF8}{gbsn}回荡\end{CJK}$\ \ $\begin{CJK}{UTF8}{gbsn}，\end{CJK}$\ \ $\begin{CJK}{UTF8}{gbsn}勾起\end{CJK}$\ \ $\begin{CJK}{UTF8}{gbsn}了\end{CJK}$\ \ $\begin{CJK}{UTF8}{gbsn}人们\end{CJK}$\ \ $\begin{CJK}{UTF8}{gbsn}对\end{CJK}$\ \ $\begin{CJK}{UTF8}{gbsn}往事\end{CJK}$\ \ $\begin{CJK}{UTF8}{gbsn}的\end{CJK}$\ \ $\begin{CJK}{UTF8}{gbsn}回忆\end{CJK}$\ \ $\begin{CJK}{UTF8}{gbsn}。\end{CJK}\\
    \\[-2.0ex] \cline{2-3} \\[-1.5ex]
    & Reference & \begin{CJK}{UTF8}{gbsn}不得\end{CJK}$\ \ $\begin{CJK}{UTF8}{gbsn}在\end{CJK}$\ \ $\begin{CJK}{UTF8}{gbsn}江河\end{CJK}$\ \ $\begin{CJK}{UTF8}{gbsn}电缆\end{CJK}$\ \ $\begin{CJK}{UTF8}{gbsn}保护区\end{CJK}$\ \ $\begin{CJK}{UTF8}{gbsn}内\end{CJK}$\ \ $\begin{CJK}{UTF8}{gbsn}抛锚\end{CJK}$\ \ $\begin{CJK}{UTF8}{gbsn}、\end{CJK}$\ \ $\begin{CJK}{UTF8}{gbsn}拖锚\end{CJK}$\ \ $\begin{CJK}{UTF8}{gbsn}、\end{CJK}$\ \ $\begin{CJK}{UTF8}{gbsn}炸鱼\end{CJK}$\ \ $\begin{CJK}{UTF8}{gbsn}、\end{CJK}$\ \ $\begin{CJK}{UTF8}{gbsn}挖沙\end{CJK}$\ \ $\begin{CJK}{UTF8}{gbsn}。\end{CJK}\\
    & Unigram DP & \begin{CJK}{UTF8}{gbsn}不得\end{CJK}$\ \ $\begin{CJK}{UTF8}{gbsn}在\end{CJK}$\ \ $\begin{CJK}{UTF8}{gbsn}江河电缆\end{CJK}$\ \ $\begin{CJK}{UTF8}{gbsn}保护\end{CJK}$\ \ $\begin{CJK}{UTF8}{gbsn}区内抛锚、\end{CJK}$\ \ $\begin{CJK}{UTF8}{gbsn}拖锚\end{CJK}$\ \ $\begin{CJK}{UTF8}{gbsn}、炸鱼、挖沙\end{CJK}$\ \ $\begin{CJK}{UTF8}{gbsn}。\end{CJK}\\
    & Bigram HDP & \begin{CJK}{UTF8}{gbsn}不得\end{CJK}$\ \ $\begin{CJK}{UTF8}{gbsn}在\end{CJK}$\ \ $\begin{CJK}{UTF8}{gbsn}江\end{CJK}$\ \ $\begin{CJK}{UTF8}{gbsn}河\end{CJK}$\ \ $\begin{CJK}{UTF8}{gbsn}电缆\end{CJK}$\ \ $\begin{CJK}{UTF8}{gbsn}保护\end{CJK}$\ \ $\begin{CJK}{UTF8}{gbsn}区内\end{CJK}$\ \ $\begin{CJK}{UTF8}{gbsn}抛\end{CJK}$\ \ $\begin{CJK}{UTF8}{gbsn}锚、拖\end{CJK}$\ \ $\begin{CJK}{UTF8}{gbsn}锚\end{CJK}$\ \ $\begin{CJK}{UTF8}{gbsn}、\end{CJK}$\ \ $\begin{CJK}{UTF8}{gbsn}炸鱼、\end{CJK}$\ \ $\begin{CJK}{UTF8}{gbsn}挖沙\end{CJK}$\ \ $\begin{CJK}{UTF8}{gbsn}。\end{CJK}\\
    & SNLM & \begin{CJK}{UTF8}{gbsn}不得\end{CJK}$\ \ $\begin{CJK}{UTF8}{gbsn}在\end{CJK}$\ \ $\begin{CJK}{UTF8}{gbsn}江河\end{CJK}$\ \ $\begin{CJK}{UTF8}{gbsn}电缆\end{CJK}$\ \ $\begin{CJK}{UTF8}{gbsn}保护区\end{CJK}$\ \ $\begin{CJK}{UTF8}{gbsn}内\end{CJK}$\ \ $\begin{CJK}{UTF8}{gbsn}抛锚\end{CJK}$\ \ $\begin{CJK}{UTF8}{gbsn}、\end{CJK}$\ \ $\begin{CJK}{UTF8}{gbsn}拖锚、\end{CJK}$\ \ $\begin{CJK}{UTF8}{gbsn}炸鱼\end{CJK}$\ \ $\begin{CJK}{UTF8}{gbsn}、\end{CJK}$\ \ $\begin{CJK}{UTF8}{gbsn}挖沙\end{CJK}$\ \ $\begin{CJK}{UTF8}{gbsn}。\end{CJK}\\
    \bottomrule[0.3ex]
    \end{tabular}
    \end{center}
        \caption{Examples of predicted segmentations on English and Chinese.\label{tb:quality}}
\end{table*}
\paragraph{Language Modeling Performance}
The above results show that the SNLM infers good word segmentations. We now turn to the question of how well it predicts held-out data. Table~\ref{tb:result-lm} summarizes the results of the language modeling experiments. Again, we see that SNLM outperforms the Bayesian models and a character LSTM. Although there are numerous extensions to LSTMs to improve language modeling performance, LSTMs remain a strong baseline~\citep{melis2017state}.

One might object that because of the lexicon, the SNLM has many more parameters than the character-level LSTM baseline model. However, unlike parameters in LSTM recurrence which are used every timestep, our memory parameters are accessed very sparsely. Furthermore, we observed that an LSTM with twice the hidden units did not improve the baseline with 512 hidden units on both phonemic and orthographic versions of Brent corpus but the lexicon could. This result suggests more hidden units are useful if the model does not have enough capacity to fit larger datasets, but that the memory structure adds other dynamics which are not captured by large recurrent networks.

\begin{table}[ht]
    \begin{center}
    \begin{tabular}{@{\hskip 2pt}l@{\hskip 2pt}c@{\hskip 2pt}c@{\hskip 3pt}c@{\hskip 3pt}c@{\hskip 3pt}c@{\hskip 3pt}}
    \toprule
     & BR-text & BR-phono & PTB & CTB & PKU\\
    \midrule
    Unigram DP & 2.33 & 2.93 & 2.25 & 6.16 & 6.88\\
    Bigram HDP & 1.96 & 2.55 & 1.80 & 5.40 & 6.42\\
    LSTM & 2.03 & 2.62 & 1.65 & 4.94 & 6.20\\
    \midrule[0.2ex]
    SNLM & \textbf{1.94} & \textbf{2.54} & \textbf{1.56} & \textbf{4.84} & \textbf{5.89}\\
    \bottomrule[0.3ex]
    \end{tabular}
    \end{center}
        \caption{Test language modeling performance (bpc).\label{tb:result-lm}}
\end{table}

\paragraph{Multimodal Word Segmentation} Finally, we discuss results on word discovery with non-linguistic context (image). Although there is much evidence that neural networks can reliably learn to exploit additional relevant context to improve language modeling performance (e.g. machine translation and image captioning), it is still unclear whether the conditioning context help to discover \emph{structure} in the data. We turn to this question here. Table~\ref{tb:result-coco} summarizes language modeling and segmentation performance of our model and a baseline character-LSTM language model on the COCO image caption dataset. We use the Elman Entropy criterion to infer the segmentation points from the baseline LM, and the MAP segmentation under our model. Again, we find our model outperforms the baseline model in terms of both language modeling and word segmentation accuracy. Interestingly, we find while conditioning on image context leads to reductions in perplexity in both models, in our model the presence of the image further improves segmentation accuracy. This suggests that our model and its learning mechanism interact with the conditional context differently than the LSTM does.

To understand what kind of improvements in segmentation performance the image context leads to, we annotated the tokens in the references with part-of-speech (POS) tags and compared relative improvements on recall between SNLM ($-$image) and SNLM ($+$image) among the five POS tags which appear more than 10,000 times. We observed improvements on ADJ ($+$4.5\%), NOUN ($+$4.1\%), VERB ($+$3.1\%). The improvements on the categories ADP ($+$0.5\%) and DET ($+$0.3\%) are were more limited. The categories where we see the largest improvement in recall correspond to those that are likely \emph{a priori} to correlate most reliably with observable features. Thus, this result is consistent with a hypothesis that the lexican is successfully acquiring knowledge about how words idiosyncratically link to visual features.

\begin{table}[ht]
    \begin{center}
    \begin{tabular}{lcccc}
    \toprule
     & bpc$\downarrow$ & P $\uparrow$ & R $\uparrow$ & F1$\uparrow$\\
    \midrule
    Unigram DP & 2.23 & 44.0 & 40.0 & 41.9\\
    Bigram HDP & 1.68 & 30.9 & 40.8 & 35.1\\
    LSTM ($-$image) & 1.55 & 31.3 & 38.2 & 34.4\\
    SNLM ($-$image) & 1.52 & 39.8 & 55.3 & 46.3\\
    \midrule[0.2ex]
    LSTM ($+$image) & 1.42 & 31.7 & 39.1 & 35.0\\
    SNLM ($+$image) & \textbf{1.38} & \textbf{46.4} & \textbf{62.0} & \textbf{53.1}\\
    \bottomrule[0.3ex]
    \end{tabular}
    \end{center}
        \caption{Language modeling (bpc) and segmentation accuracy on COCO dataset. $+$image indicates that the model has access to image context. \label{tb:result-coco}}
    \vskip-2.0ex
\end{table}

\paragraph{Segmentation State-of-the-Art}
The results reported are not the best-reported numbers on the English phoneme or Chinese segmentation tasks. As we discussed in the introduction, previous work has focused on segmentation in isolation from language modeling performance. Models that obtain better segmentations include the adaptor grammars (F1: 87.0) of \citet{johnson2009improving} and the feature-unigram model (88.0) of \citet{berg2010painless}. While these results are better in terms of segmentation, they are weak language models (the feature unigram model is effectively a unigram word model; the adaptor grammar model is effectively phrasal unigram model; both are incapable of generalizing about substantially non-local dependencies). 
Additionally, the features and grammars used in prior work reflect certain English-specific design considerations (e.g., syllable structure in the case of adaptor grammars and phonotactic equivalence classes in the feature unigram model), which make them questionable models if the goal is to explore what models and biases enable word discovery in general.
For Chinese, the best nonparametric models perform better at segmentation~\citep{zhao:2008,mochihashi2009bayesian}, but again they are weaker language models than neural models. The neural model of \citet{sun2018unsupervised} is similar to our model without lexical memory or length regularization; it obtains 80.2 F1 on the PKU dataset; however, it uses gold segmentation data during training and hyperparameter selection,\footnote{\url{https://github.com/Edward-Sun/SLM/blob/d37ad735a7b1d5af430b96677c2ecf37a65f59b7/codes/run.py##L329}} whereas our approach requires no gold standard segmentation data.

\section{Related Work}
Learning to discover and represent temporally extended structures in a sequence is a fundamental problem in many fields. For example in language processing, unsupervised learning of multiple levels of linguistic structures such as morphemes~\citep{snyder2008unsupervised}, words~\citep{goldwater2009bayesian,mochihashi2009bayesian,wang2014empirical} and phrases~\citep{klein2001distributional} have been investigated. Recently, speech recognition has benefited from techniques that enable the discovery of subword units~\citep{chan:2017,wang:2017}; however, in that work, the optimally discovered character sequences look quite unlike orthographic words. In fact, the model proposed by \citet{wang:2017} is essentially our model without a lexicon or the expected length regularization, i.e., ($-$memory, $-$length), which we have shown performs quite poorly in terms of segmentation accuracy. Finally, some prior work has also sought to discover lexical units directly from speech based on speech-internal statistical regularities~\citep{kamper:2016}, as well as jointly with grounding~\citep{chrupala2017representations}.

\section{Conclusion}
Word discovery is a fundamental problem in language acquisition. While work studying the problem in isolation has provided valuable insights (showing both what data is sufficient for word discovery with which models), this paper shows that neural models offer the flexibility and performance to productively study the various facets of the problem in a more unified model.
While this work unifies several components that had previously been studied in isolation, our model assumes access to phonetic categories. The development of these categories likely interact with the development of the lexicon and acquisition of semantics~\citep{feldman:2013,fourtassi:2014}, and thus subsequent work should seek to unify more aspects of the acquisition problem.

\subsubsection*{Acknowledgments}
We thank Mark Johnson, Sharon Goldwater, and Emmanuel Dupoux, as well as many colleagues at DeepMind, for their insightful comments and suggestions for improving this work and the resulting paper.

\bibliography{acl2019}
\bibliographystyle{acl_natbib}

\clearpage

\appendix
\begin{table*}[!htb]
    \begin{center}\scriptsize
    \begin{tabular}{lrrrrrrrrrrrrrrrr}
    \toprule
      & \multicolumn{3}{c}{Sentence} & \multicolumn{3}{c}{Char. Types} & \multicolumn{3}{c}{Word Types} & \multicolumn{3}{c}{Characters} & \multicolumn{3}{c}{Average Word Length}\\
    \midrule
     & Train & Valid & Test & Train & Valid & Test & Train & Valid & Test & Train & Valid & Test & Train & Valid & Test\\
    \midrule
    BR-text & 7832 & 979 & 979 & 30 & 30 & 29 & 1237 & 473 & 475 & 129k & 16k & 16k & 3.82 & 4.06 & 3.83\\
    BR-phono & 7832 & 978 & 978 & 51 & 51 & 50 & 1183 & 457 & 462 & 104k & 13k & 13k & 2.86 & 2.97 & 2.83\\
    PTB & 42068 & 3370 & 3761 & 50 & 50 & 48 & 10000 & 6022 & 6049 & 5.1M & 400k & 450k & 4.44 & 4.37 & 4.41\\
    CTB & 50734 & 349 & 345 & 160 & 76 & 76 & 60095 & 1769 & 1810 & 3.1M & 18k & 22k & 4.84 & 5.07 & 5.14\\
    PKU & 17149 & 1841 & 1790 & 90 & 84 & 87 & 52539 & 13103 & 11665 & 2.6M & 247k & 241k & 4.93 & 4.94 & 4.85\\
    COCO & 8000 & 2000 & 10000 & 50 & 42 & 48 & 4390 & 2260 & 5072 & 417k & 104k & 520k & 4.00 & 3.99 & 3.99\\
    \bottomrule
    \end{tabular}
    \end{center}
    \caption{Summary of Dataset Statistics. \label{tb:dataset}}
\end{table*}

\section{Dataset statistics}\label{sec:app-stats}
Table~\ref{tb:dataset} summarizes dataset statistics.

\section{Image Caption Dataset Construction}\label{sec:app-caption}
We use 8000, 2000 and 10000 images for train, development and test set in order of integer ids specifying image in cocoapi\footnote{https://github.com/cocodataset/cocoapi} and use first annotation provided for each image. We will make pairs of image id and annotation id available from \url{https://s3.eu-west-2.amazonaws.com/k-kawakami/seg.zip}.

\section{SNLM Model Configuration}\label{sec:config}
For each RNN based model we used 512 dimensions for the character embeddings and the LSTMs have 512 hidden units. All the parameters, including character projection parameters, are randomly sampled from uniform distribution from $-0.08$ to $0.08$. The initial hidden and memory state of the LSTMs are initialized with zero. A dropout rate of 0.5 was used for all but the recurrent connections.

To restrict the size of memory, we stored substrings which appeared $F$-times in the training corpora and tuned $F$ with grid search. The maximum length of subsequences $L$ was tuned on the held-out likelihood using a grid search. Tab.~\ref{tb:params} summarizes the parameters for each dataset. Note that we did not tune the hyperparameters on segmentation quality to ensure that the models are trained in a purely unsupervised manner assuming no reference segmentations are available.

\begin{table*}[ht]
    \begin{center}%
    \begin{tabular}{l@{\hskip 9.5pt}c@{\hskip 9.5pt}c@{\hskip 9.5pt}c}
    \toprule
     & max len (L) & min freq (F) & $\lambda$\\
    \midrule
    BR-text  & 10 & 10 & 7.5e-4\\
    BR-phono & 10 & 10 & 9.5e-4\\
    PTB      & 10 & 100 & 5.0e-5\\
    CTB      & 5 & 25  & 1.0e-2\\
    PKU      & 5 & 25  & 9.0e-3\\
    COCO     & 10 & 100 & 2.0e-4\\
    \bottomrule[0.3ex]
    \end{tabular}
    \end{center}
    \caption{Hyperparameter values used.\label{tb:params}}
\end{table*}

\section{DP/HDP Inference}
\label{app:inference}
By integrating out the draws from the DP's, it is possible to do inference using Gibbs sampling directly in the space of segmentation decisions. We use 1,000 iterations with annealing to find an approximation of the MAP segmentation and then use the corresponding posterior predictive distribution to estimate the held-out likelihood assigned by the model, marginalizing the segmentations using appropriate dynamic programs. The evaluated segmentation was the most probable segmentation according to the posterior predictive distribution.

In the original Bayesian segmentation work, the hyperparameters (i.e., $\alpha_0$, $\alpha_1$, and the components of $p_0$) were selected subjectively. To make comparison with our neural models fairer, we instead used an empirical approach and set them using the held-out likelihood of the validation set. However, since this disadvantages the DP/HDP models in terms of segmentation, we also report the original results on the BR corpora.

\section{Learning}\label{sec:learning}
The models were trained with the Adam update rule~\citep{kingma2014adam} with a learning rate of 0.01. The learning rate is divided by 4 if there is no improvement on development data. The maximum norm of the gradients was clipped at 1.0.

\section{Evaluation Metrics}\label{sec:metrics}
\paragraph{Language Modeling} We evaluated our models with bits-per-character (bpc), a standard evaluation metric for character-level language models. Following the definition in \citet{graves2013generating}, bits-per-character is the average value of $-\log_{2} p(x_{t}\mid \boldsymbol{x}_{<t})$ over the whole test set,
\begin{align*}
    \textit{bpc} = -\frac{1}{|\boldsymbol{x}|} \log_{2} p(\boldsymbol{x}),
\end{align*}
where $|\boldsymbol{x}|$ is the length of the corpus in characters. The bpc is reported on the test set.

\paragraph{Segmentation} We also evaluated segmentation quality in terms of precision, recall, and F1 of word tokens~\citep{brent1999efficient,venkataraman2001statistical,goldwater2009bayesian}. To get credit for a word, the models must correctly identify both the left and right boundaries. For example, if there is a pair of a reference segmentation and a prediction,
\begin{align*}
    \textrm{Reference: }&\texttt{do you see a boy} \\
    \textrm{Prediction: }&\texttt{doyou see a boy}
\end{align*}
then 4 words are discovered in the prediction where the reference has 5 words. 3 words in the prediction match with the reference. In this case, we report scores as precision = 75.0 (3/4), recall = 60.0 (3/5), and F1, the harmonic mean of precision and recall, 66.7 (2/3). To facilitate comparison with previous work, segmentation results are reported on the union of the training, validation, and test sets.
\end{document}